\documentclass[10pt,twocolumn,letterpaper]{article}

\usepackage[pagenumbers]{wacv} % To force page numbers, e.g. for an arXiv version

\usepackage[accsupp]{axessibility} % Improves PDF readability for those with disabilities.

\usepackage{graphicx}
\usepackage{amsmath}
\usepackage{amssymb}
\usepackage{booktabs}

\usepackage[dvipsnames]{xcolor}
\usepackage{multirow}
\usepackage{colortbl}
\usepackage{comment}
\usepackage{times}
\usepackage{epsfig}
\usepackage{xcolor}
\usepackage{boldline}
\usepackage{algpseudocode}
\usepackage{algorithm}

\usepackage[pagebackref,breaklinks,colorlinks]{hyperref}

\usepackage[capitalize]{cleveref}
\crefname{section}{Sec.}{Secs.}
\Crefname{section}{Section}{Sections}
\Crefname{table}{Table}{Tables}
\crefname{table}{Tab.}{Tabs.}

\begin{document}

\title{RapidNet: Multi-Level Dilated Convolution Based Mobile Backbone}

\author{Mustafa Munir\\
The University of Texas at Austin\\
{\tt\small mmunir@utexas.edu}
\and
Md Mostafijur Rahman\\
The University of Texas at Austin\\
{\tt\small mostafijur.rahman@utexas.edu }
\and
Radu Marculescu\\
The University of Texas at Austin\\
{\tt\small radum@utexas.edu} \\
}

\maketitle

\def\thefootnote{$^1$}
\footnotetext{\scriptsize Code: \url{https://github.com/mmunir127/RapidNet-Official}}

%%%%%%%%% ABSTRACT
\begin{abstract}
Vision transformers (ViTs) have dominated computer vision in recent years. However, ViTs are computationally expensive and not well suited for mobile devices; this led to the prevalence of convolutional neural network (CNN) and ViT-based hybrid models for mobile vision applications. Recently, Vision GNN (ViG) and CNN hybrid models have also been proposed for mobile vision tasks. However, all of these methods remain slower compared to pure CNN-based models. In this work, we propose Multi-Level Dilated Convolutions to devise a purely CNN-based mobile backbone. Using Multi-Level Dilated Convolutions allows for a larger theoretical receptive field than standard convolutions. Different levels of dilation also allow for interactions between the short-range and long-range features in an image. Experiments show that our proposed model outperforms state-of-the-art (SOTA) mobile CNN, ViT, ViG, and hybrid architectures in terms of accuracy and/or speed on image classification, object detection, instance segmentation, and semantic segmentation. Our fastest model, RapidNet-Ti, achieves 76.3\% top-1 accuracy on ImageNet-1K with 0.9 ms inference latency on an iPhone 13 mini NPU, which is faster and more accurate than MobileNetV2x1.4 (74.7\% top-1 with 1.0 ms latency). Our work shows that pure CNN architectures can beat SOTA hybrid and ViT models in terms of accuracy and speed when designed properly$^1$.

\end{abstract}

\vspace{-6mm}

%%%%%%%%% BODY TEXT
\section{Introduction}
\label{sec:intro}

The field of deep learning has witnessed remarkable advancements in computer vision in the last decade \cite{munir2024three}, from image classification and object detection to generative vision tasks, such as image synthesis \cite{huang2018imsynthesis, brock2018large,  dhariwal2021diffusion} and video synthesis \cite{esser2023structure, liu2021generative} using generative adversarial networks (GANs) \cite{goodfellow2014generative} and diffusion models \cite{ho2020denoising}. This evolution has been fueled by diverse architectural paradigms, including Convolutional Neural Networks (CNNs) \cite{lecun1998gradient, Resnet, MobileNet, Densenet, liu2022convnet}, Vision Transformers (ViTs) \cite{ViT, liu2021swin, carion2020end}, and Multi-Layer Perceptron (MLP)-based \cite{resmlp, mlpmixer} models. CNNs and MLPs interpret images as pixel grids, while ViTs represent them as sequences of patches \cite{ViT}, enabling them to be processed by transformers \cite{vaswani2017attention}. ViTs also have global receptive fields and capture distant interactions within images, unlike CNNs which have local receptive fields \cite{ViT}.

The emergence of Vision Graph Neural Networks (ViGs), exemplified by models like ViG \cite{Vision_GNN}, ViHGNN \cite{ViHGNN}, and MobileViG \cite{MobileViG, MobileViGv2}, introduced graph-based approaches, which connect image patches through graph structures. While ViG-based models demonstrate their potential in capturing global object interactions, they incur large computational costs due to graph construction \cite{GreedyViG}.

The demand for deploying powerful AI applications directly on mobile devices has led to the exploration of lightweight models \cite{MobileNet, MobileNetv2, MobileFormer, MetaFormer}. Early efforts with CNNs on mobile platforms paved the way for hybrid CNN-ViT architectures, but the computational cost of the self-attention operation in ViTs poses significant challenges for mobile applications \cite{MobileViT, MobileViTv2}. The exploration of ViTs, ViGs, and hybrid architectures for mobile devices has led to many advances in accuracy, but state-of-the-art (SOTA) results can still be achieved using only CNN-based models.

One avenue to make CNN-based models competitive is dilated convolutions \cite{yu2015multi}. Dilated convolutions can increase the receptive field of a convolution operation, but with a lower cost than increasing the kernel size. This is because a dilated convolution effectively increases the kernel's receptive field by inserting "gaps" in between elements of the kernel by a dilation factor \cite{yu2015multi}. This means that a 3 $\times$ 3 convolution with a dilation factor of 2 will have a receptive field equal to a 5 $\times$ 5 convolution while using less parameters. Thus, we can use dilated convolutions for increasing the receptive field in our network, similar to how hybrid CNN-ViT and CNN-ViG architectures use ViTs and ViGs to increase their receptive field.

%-------------------------------------------------------------------------

%------------------------------------------------------------------------- 
\begin{figure*}[t]
\centering
\includegraphics[width=\linewidth,]{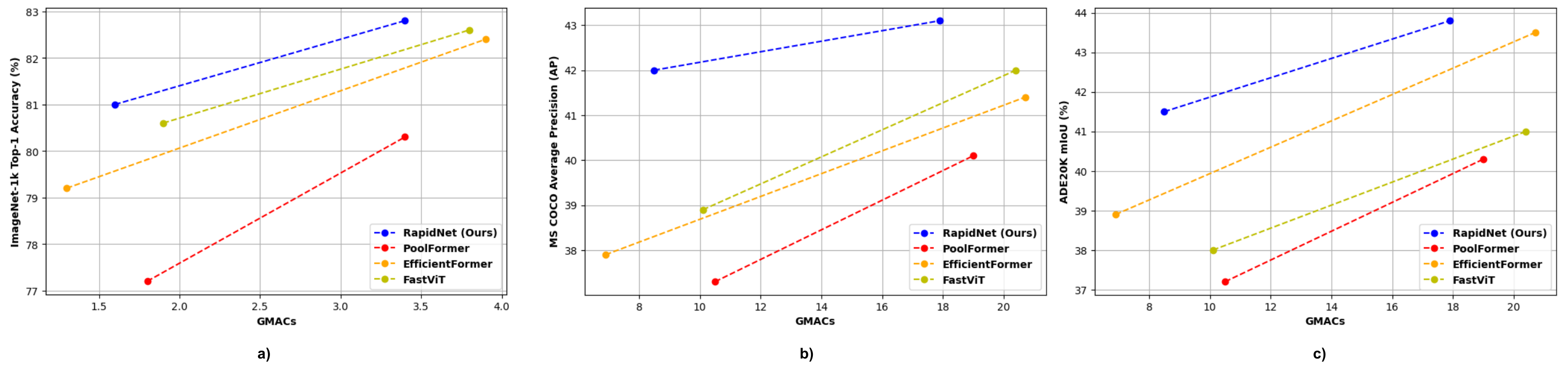}
\caption{\textbf{Comparison of accuracy on ImageNet-1K, Average Precision (AP) on MS COCO, and mean Intersection over Union (mIoU) on ADE20K}. a) RapidNet achieves the best accuracy-GMACs tradeoff on all model sizes compared. b) RapidNet achieves the best AP-GMACs tradeoff on all model sizes compared. c) RapidNet achieves the best mIoU-GMACs tradeoff on all model sizes compared. GMACs are computed using a resolution of 224 $\times$ 224 for a) and a resolution of 512 $\times$ 512 for b) and c).}
\label{fig:pareto}
\end{figure*}

%------------------------------------------------------------------------- 

 In this work, we propose Multi-Level Dilated Convolution blocks to create a CNN-based architecture competitive with SOTA CNN, ViT, ViG, and hybrid models. Our newly introduced architecture, RapidNet, is faster, less computationally expensive in terms of GMACS, and/or more accurate compared to other mobile architectures as shown in Figure \ref{fig:pareto}. Indeed, our experimental results show that our proposed RapidNet architecture outperforms competing SOTA models across all model sizes for the tasks of image classification, object detection, and semantic segmentation. We summarize our contributions as follows:

\begin{enumerate}
    \item We propose using Multi-Level Dilated Convolutions (MLDC) to enable processing features at different levels of dilation in parallel. MLDC expands the receptive field of convolutions, thus allowing for an efficient CNN-based alternative to ViG and ViT-based models.
    \item We propose a novel efficient CNN-based architecture, RapidNet, which uses MLDC, reparameterizable large kernel depthwise convolutions \cite{CPE, FastViT}, and a large kernel feedforward network (FFN) \cite{FastViT}.
    \item We conduct comprehensive experiments to demonstrate the effectiveness of the RapidNet architecture, which beats the existing efficient ViG, CNN, and ViT architectures in terms of top-1 accuracy, GMACs, and/or latency on ImageNet-1k \cite{imagenet1k} image classification, COCO \cite{coco} object detection, COCO \cite{coco} instance segmentation, and ADE20K \cite{ADE20K} semantic segmentation. Specifically, our RapidNet-M model achieves a top-1 accuracy of 81.0\% on ImageNet classification, 42.0 Average Precision (AP) on COCO object detection, and 41.5 mean Intersection over Union (mIoU) on ADE20K semantic segmentation.
\end{enumerate}

The paper is organized as follows. Section \ref{sec:rel_work} covers related work on dilated convolutions and efficient computer vision. Section \ref{sec:architecture} describes Multi-Level Dilated Convolutions, the usage of a large kernel FFN, our RapidNet architecture, and the network configuration of RapidNet for different model sizes. Section \ref{sec:results} describes our experimental setup and results for ImageNet-1k image classification, COCO object detection, COCO instance segmentation, and ADE20K semantic segmentation. Lastly, Section \ref{sec:conc} summarizes our main contributions.

\section{Related Work}
\label{sec:rel_work}

In this section, we review dilated convolutions and previous work in the mobile computer vision space.

\subsection{Dilated Convolution}

Dilated convolutions \cite{holschneider1990real} introduce "gaps" between kernel elements by a dilation factor. The "gaps" are introduced by inserting zeros between each pixel in the convolutional kernel \cite{yu2015multi}, thus allowing for an expanded receptive field without increasing the number of parameters \cite{yu2015multi, mehta2018espnet}.

\begin{figure}[h]
\centering
\includegraphics[width=\columnwidth]{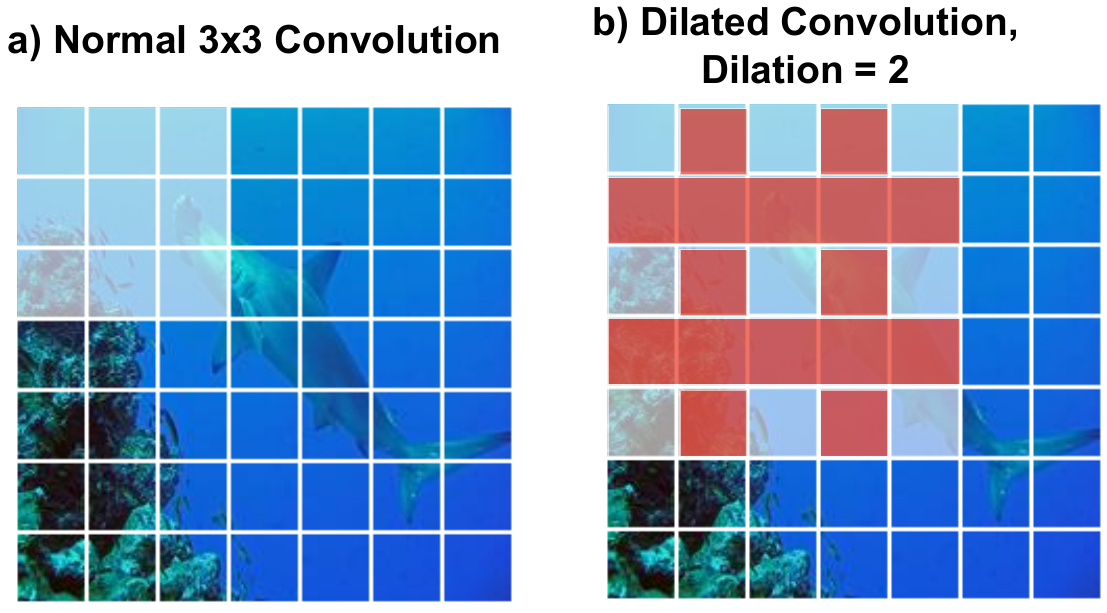}
\caption{\textbf{Comparison of regular and dilated convolution.} a) In a regular convolution with a kernel size of 3 in a 7 $\times$ 7 image, we can see the convolution is applied to the 3 $\times$ 3 patches in the grid. b) In a dilated convolution with a kernel size of 3 and a dilation factor of 2 in a 7 $\times$ 7 image, we can see the convolution is applied to the 5 $\times$ 5 patches in the grid thereby expanding the receptive field. This is done by skipping the patches in red in the convolution, preserving the parameters needed for 3 $\times$ 3 convolution, but expanding the receptive field to that of a 5 $\times$ 5 convolution.} 
\label{fig:Dilated}
\end{figure}

The expanded receptive field enables the model to capture a broader range of contextual information, thus facilitating more effective feature extraction. Dilated convolutions enhance the model's ability to capture long-range dependencies and improve its overall performance in tasks such as semantic segmentation \cite{yu2015multi, mehta2018espnet}.

In Figure \ref{fig:Dilated}, we show how a  3 $\times$ 3 dilated convolution and 3 $\times$ 3 regular convolution differ as the dilated convolution introduces "gaps" in between the kernel elements. In Figure \ref{fig:Dilated}a we can see the theoretical receptive field of the 3 $\times$ 3 convolution is 3 $\times$ 3, while in Figure \ref{fig:Dilated}b we see that the theoretical receptive field of the 3$ \times$ 3 dilated convolution with dilation factor of 2 is 5 $\times$ 5. For a $k \times k$ dilated convolution with a dilation factor of $d$, the theoretical receptive field (TRF) of the kernel is:

\begin{equation}
\tag{1}
TRF = (((k-1) \times d) + 1) \times (((k-1) \times d) + 1) 
\end{equation}

where $d-1$ is the number of "gaps" between pixels; thus, for a regular convolution, $d = 1$. Dilated convolutions decrease computational cost as only $k \times k$ pixels participate in the convolution even though theoretical receptive field of the convolution is increased \cite{mehta2018espnet}.

Past work on large kernel convolutions and dilated convolutions have shown the effectiveness of increasing the receptive field of a convolutional filter \cite{31_31_Conv, liu2022convnet}. Large kernel convolutions are computationally expensive, thus one way to decrease the computation needed for increasing the receptive field is to use dilated convolutions. ESPNet \cite{mehta2018espnet} employs pointwise convolutions to reduce the computational cost of dilated convolutions and uses dilated convolutions to learn representations from the larger receptive field.

\subsection{Mobile Vision}

We can break up the past approaches in the mobile vision space into CNN-based approaches, CNN-ViT approaches, and CNN-ViG approaches.

\textbf{1. CNN-Based Approaches}
In the domain of mobile vision, CNNs have historically been the mainstream architecture, with notable contributions from MobileNets \cite{MobileNet, MobileNetv2}, EfficientNets \cite{tan2019efficientnet, tan2021efficientnetv2}, ShuffleNets \cite{shufflenet, shufflenetv2}, and SqueezeNet \cite{SqueezeNet}. MobileNet \cite{MobileNet} introduced depthwise separable convolutions, achieving comparable performance to standard convolutions with significantly lower computational cost by splitting a full convolution into a factorized version using a depthwise convolution and pointwise convolution. MobileNetv2 \cite{MobileNetv2} enhanced this with the introduction of inverted residuals and linear bottlenecks. EfficientNet \cite{tan2019efficientnet, tan2021efficientnetv2} leveraged neural architecture search to produce fast and accurate models. ShuffleNet introduced pointwise group convolution and channel shuffle \cite{shufflenet}. SqueezeNet helped push smaller networks with model compression achieving AlexNet level accuracy with 50$\times$ fewer parameters \cite{SqueezeNet}.

\textbf{2. CNN-ViT-Based Approaches}
Recent advancements in the efficient computer vision space have led to the emergence of hybrid CNN-ViT-based models, particularly focusing on high accuracy while reducing the latency of the self-attention operation. The EfficientFormer family of models \cite{EfficientFormer, EfficientFormerv2} combine local processing using CNNs with multi-head self-attention (MHSA) operations for global processing. MobileViT and MobileViTv2 \cite{MobileViT, MobileViTv2} are also notable examples of such hybrid models that combine MobileNetv2 \cite{MobileNetv2} blocks and MHSA blocks, aiming to effectively capture both local and global information.

\textbf{3. CNN-ViG-Based Approaches}
Graph Neural Networks (GNNs) have traditionally been used in research on biological, social, and citation networks, but have grown in usage in the computer vision domain \cite{Vision_GNN} too. Vision GNN (ViG) \cite{Vision_GNN} used GNNs as a general-purpose vision backbone by splitting an image into patches and connecting the patches based on the K-Nearest Neighbors (KNN) algorithm. MobileViG \cite{MobileViG} introduces a hybrid CNN-GNN architecture, utilizing Sparse Vision Graph Attention (SVGA) for efficient graph construction to make ViG fast on mobile devices. While MobileViG achieves high accuracy and low latency, it limits its usage of graph convolutions to the lowest resolution stage to decrease the impact on latency.

\section{Method}
\label{sec:architecture}

%------------------------------------------------------------------------- 
\begin{figure*}[t]
\centering
\includegraphics[width=0.85\linewidth]{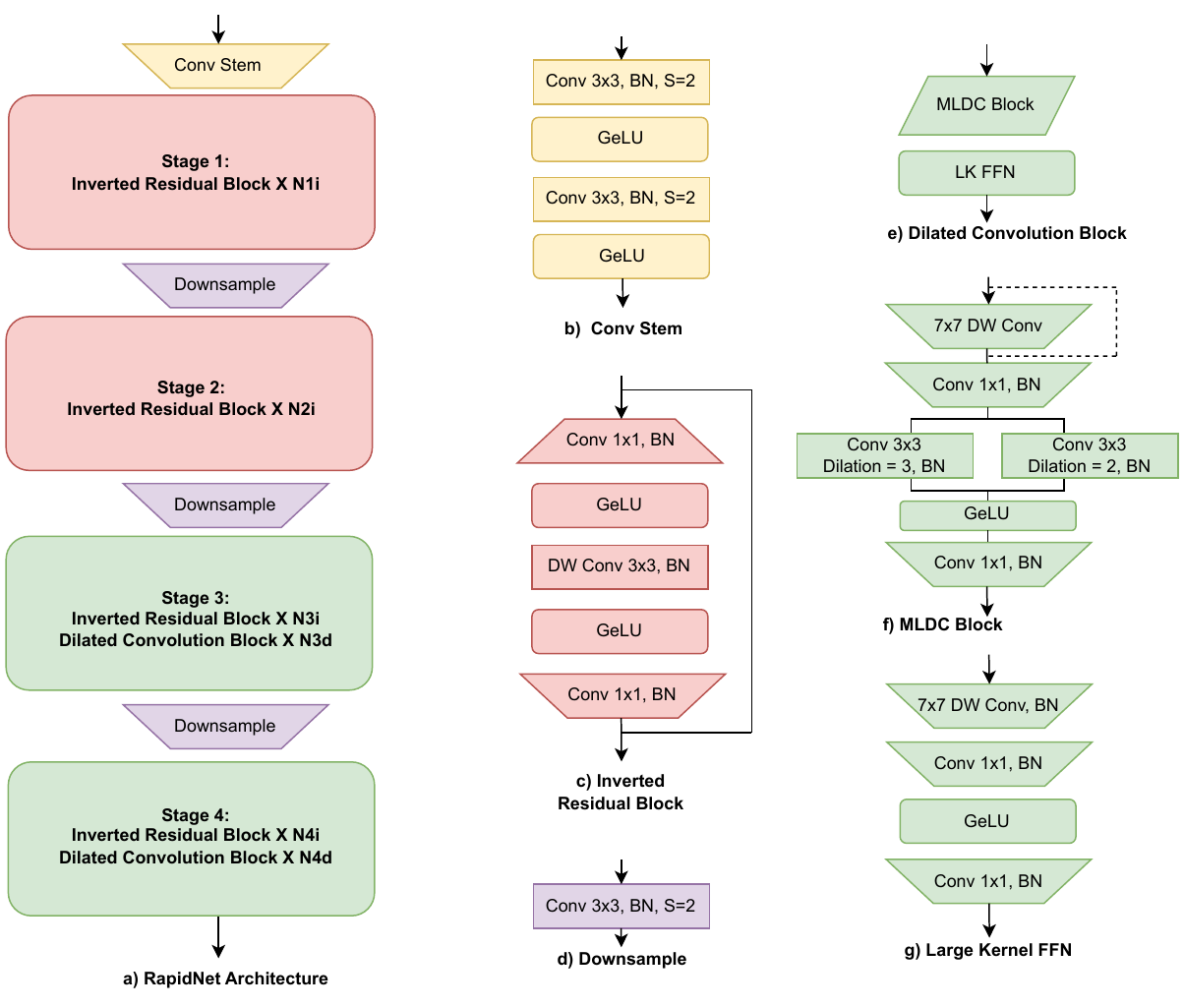}
\caption{\textbf{RapidNet architecture.} (a) Network architecture showing the stages and layers, where N1i, N2i, N3i, N4i, N3d, and N4d represent the number of Inverted Residual Blocks and Dilated Convolution Blocks in the RapidNet-Ti, S, M, and B configurations and S represents the stride of the convolutions. (b) The Conv Stem. (c) Inverted Residual Block. (d) Downsample. (e) Dilated Convolution Block. (f) Multi-Level Dilated Convolution (MLDC) Block. (g) Large Kernel FFN.}
\label{fig:RapidNet_Architecture}
\end{figure*}
%------------------------------------------------------------------------- 

In this section, we describe how we use Multi-Level Dilated Convolutions and provide details on the RapidNet architecture design. More precisely, Section 3.1 describes why we use dilated convolutions, Section 3.2 describes Multi-Level Dilated Convolutions. Section 3.3 describes our usage of a large kernel FFN. Section 3.4 describes how we combine the MLDC blocks, large kernel FFN, and inverted residual blocks for local processing to create the RapidNet architecture. Lastly, Section 3.5 describes our RapidNet network architecture for different model sizes.

\subsection{Why Dilated Convolutions?}

MobileViG \cite{MobileViG} leverages graph convolution to perform global processing in its lowest resolution stage. However, if the cost of dilated convolution is no greater than the cost of graph convolution, then can we achieve better performance through the use of dilated convolutions? In MobileViG \cite{MobileViG}, the authors propose a static graph construction method called Sparse Vision Graph Attention (SVGA) to connect to every $K^{th}$ pixel in the row and column of the graph. Since SVGA and graph convolution are only used in the lowest resolution stage of MobileViG \cite{MobileViG} and $K=2$ in the MobileViG implementation, the graph convolution would have an theoretical receptive field over all of the patches in the image. Since the input image resolution is halved in the stem and in each downsample layer, the theoretical receptive field for SVGA in the final stage of MobileViG is:

\begin{equation}
\tag{2}
TRF_{MobileViG} = 7\times7
\end{equation}

If we replace the graph convolution of MobileViG \cite{MobileViG} with a 3 $\times$ 3 convolution and a dilation of 3 then we can achieve the same theoretical receptive field as MobileViG:

\begin{equation}
\tag{3}
TRF_{Kernel = 3\times3, Dilation = 3} = 7\times7
\end{equation}

Using 3 $\times$ 3 convolutions has been shown to be effective in past works \cite{wang2023repvit, ding2021repvgg} and using dilated 3 $\times$ 3 convolutions can achieve a larger theoretical receptive field, without a major hit in latency. Two dilated convolutions at different levels of dilation can be used to process features with different theoretical receptive fields in parallel. We used dilated convolutions instead of deformable convolutions due to them being better suited for our aim of computational efficiency, due to no additional learnable parameters \cite{dai2017deformable}. We also provide an ablation study in the Supplementary Materials Table \ref{tab:ablation_kernel_dilation_deformable}, which shows our better performance using our Multi-Level Dilated Convolution.

\subsection{Multi-Level Dilated Convolution}
 
We propose using two parallel dilated convolutions in place of the max-relative graph convolution \cite{maxrel, MobileViG} used in MobileViG. Both of the dilated convolutions are at different levels of dilation, where dilated convolution one has a dilation factor of 2 and dilated convolution two has a dilation factor of 3. Since we use 3 $\times$ 3 convolutions instead of the pointwise convolutions of MobileViG \cite{MobileViG}, we can get a theoretical receptive field of a 5 $\times$ 5 convolution and a 7 $\times$ 7 convolution for our dilated convolutions with the parameter cost of a 3 $\times$ 3 convolution for each dilated convolution. 

We note that both convolutions occur in parallel and we sum their output after batch normalization (BN) \cite{BatchNorm} and GeLU activation \cite{GeLU}. These parallel dilated convolutions at different dilation levels allow us to process features in different receptive fields, thus better enabling feature extraction. The usage of dilated convolutions as opposed to large kernel convolutions is to decrease the computational cost from processing via a larger receptive field.

In our RapidNet architecture, the Dilated Convolution Block consists of our MLDC Block followed by a large kernel FFN \cite{FastViT}. The MLDC Block shown in Figure \ref{fig:RapidNet_Architecture}f consists of a 7 $\times$ 7 depthwise convolution \cite{CPE, FastViT} with a reparameterizable skip connection. Given an input feature $X\in\mathbb{R}^{N \times N}$, during training this can be expressed as: 

\begin{equation}
\tag{4}
Y=(X + DW_{7\times7}(X))
\end{equation}

During inference this can be expressed as: 

\begin{equation}
\tag{5}
Y=DW_{7\times7}(X)
\end{equation}

where $Y\in\mathbb{R}^{N \times N}$ and $DW_{7\times7}$ is a 7 $\times$ 7 depthwise convolution. This is followed by a pointwise convolution and BN, then two dilated convolution blocks expressed as:

\begin{equation}
\tag{6}
Z = \sigma(Dilated_2(Y) + Dilated_3(Y))
\end{equation}

where $Z\in\mathbb{R}^{N \times N}$, $Dilated_{2}$ and $Dilated_{3}$ are 3 $\times$ 3 kernel convolutions with dilation factors of 2 and 3 respectively, and $\sigma$ is a GeLU activation.

Following the MLDC Block, we use the large kernel FFN module as used in \cite{FastViT}, which can be seen in Figure \ref{fig:RapidNet_Architecture}g. The large kernel FFN module is a 7 $\times$ 7 depthwise convolution followed by a two layer MLP expressed as:

\begin{equation}
\tag{7}
Out= BN (FC_{2}(\sigma (FC_{1}(BN(DW_{7\times7}(Z))))))
\end{equation}
where $Out\in\mathbb{R}^{N \times N}$, $FC_{1}$ and $FC_{2}$ are fully connected layers, $\sigma$ is once again GeLU, $BN$ is batch normalization, and $DW_{7\times7}$ is again a 7 $\times$ 7 depthwise convolution. We call this combination of the MLDC Block and large kernel FFN the Dilated Convolution block, as shown in Figure \ref{fig:RapidNet_Architecture}e.

A reparameterizable 7 $\times$ 7 depthwise convolution is introduced into the MLDC block to also expand the receptive field \cite{CPE}. The reparameterization follows the method of \cite{FastViT} to eliminate a skip connection at inference time thereby decreasing latency without damaging accuracy.

\subsection{Large Kernel FFN}

Since we do not use self-attention token mixers or the graph-based mixing of MobileViG \cite{MobileViG}, we need to employ another efficient method to expand our theoretical receptive field. For the last two stages in our architecture, we do this through our MLDC Block, but using the MLDC Block in all four stages is too computationally expensive. Thus, an efficient approach to improve the receptive field of our architecture in the first two stages is to incorporate depthwise large kernel convolutions \cite{FastViT} in the FFN. Following the method of \cite{FastViT}, we incorporate depthwise 7 $\times$ 7 kernel convolutions in the FFN. The architecture of the large kernel FFN (LK FFN) is shown in Figure \ref{fig:RapidNet_Architecture}g. The LK FFN block is similar to past works \cite{liu2022convnet, MobileViG}, but utilizes large kernel convolutions to enhance the receptive field and bolster model robustness as shown in \cite{wang2022can}. Convolutional FFN blocks have been shown to exhibit greater robustness compared to standard FFN blocks \cite{FastViT} as shown in \cite{mao2022towards}. Thus, inspired by \cite{FastViT} we integrate large kernel depthwise convolutions into our FFN as an effective method for elevating model performance and robustness while minimizing the impact on latency.

\subsection{RapidNet Architecture}
\begin{table*}[h]
\small
%\fontsize{10}{11}\selectfont %\small % or \footnotesize, \scriptsize, etc.
\caption{\textbf{Architecture details of RapidNet} showing configuration of the stem, stages, output size, downsample layers, and classification head. \textit{Channels} represents the channel width. \textit{IRB} represents the Inverted Residual Block. \textit{DCB} represents the Dilated Convolution Block. Lin. MLP stands for linear MLP. } 
\centering
\setlength{\tabcolsep}{4pt}
\begin{tabular}{|c|c|c|c|c|c|}
\hline
Stage                           & Output Size & RapidNet-Ti & RapidNet-S & RapidNet-M & RapidNet-B \\ \hline \rule{0pt}{4ex}
Stem                         & $\dfrac{H}{4} \times \dfrac{W}{4}$       & Conv $\times$2, Stride=2         & Conv $\times$2, Stride=2             & Conv $\times$2, Stride=2             & Conv $\times$2, Stride=2            \\[8pt] \hline \rule{0pt}{4ex}
Stage 1          & $\dfrac{H}{4} \times \dfrac{W}{4}$   & $ \begin{array}{ccc} IRB \times2 \\ Channels = 32 \end{array} $             & $ \begin{array}{ccc} IRB \times3 \\ Channels = 32 \end{array} $    & $ \begin{array}{ccc} IRB \times3 \\ Channels = 32 \end{array} $          
& $ \begin{array}{ccc} IRB \times3 \\ Channels = 64 \end{array} $ 
 \\[8pt] \hline \rule{0pt}{4ex}     
 Downsample               & $\dfrac{H}{8} \times \dfrac{W}{8}$                  & Conv, Stride=2              & Conv, Stride=2   & Conv, Stride=2             & Conv, Stride=2             \\[8pt] \hline \rule{0pt}{4ex}
Stage 2           & $\dfrac{H}{8} \times \dfrac{W}{8}$   & $ \begin{array}{ccc} IRB \times2 \\ Channels = 64 \end{array} $           & $ \begin{array}{ccc} IRB \times3 \\ Channels = 64 \end{array} $     & $ \begin{array}{ccc} IRB \times3 \\ Channels = 64 \end{array} $          
& $ \begin{array}{ccc} IRB \times3 \\ Channels = 128 \end{array} $
\\[8pt] \hline \rule{0pt}{4ex}

 Downsample               & $\dfrac{H}{16} \times \dfrac{W}{16}$                  & Conv, Stride=2              & Conv, Stride=2    & Conv, Stride=2             & Conv, Stride=2             \\[8pt] \hline \rule{0pt}{4ex}
 
Stage 3              & $\dfrac{H}{16} \times \dfrac{W}{16}$  & $ \begin{array}{ccc} IRB \times6 \\ DCB \times2 \\ Channels = 112 \end{array} $             & $ \begin{array}{ccc} IRB \times9 \\ DCB \times3 \\ Channels = 112 \end{array} $          
 & $ \begin{array}{ccc} IRB \times9 \\ DCB \times3 \\ Channels = 160 \end{array} $  & $ \begin{array}{ccc} IRB \times9 \\ DCB \times3 \\ Channels = 224 \end{array} $
\\[8pt] \hline \rule{0pt}{4ex}

 Downsample               & $\dfrac{H}{32} \times \dfrac{W}{32}$                  & Conv, Stride=2              & Conv, Stride=2    & Conv, Stride=2             & Conv, Stride=2             \\[8pt] \hline \rule{0pt}{4ex}
 
Stage 4                  & $\dfrac{H}{32} \times \dfrac{W}{32}$   & $ \begin{array}{ccc} IRB \times2 \\ DCB \times2 \\ Channels = 224 \end{array} $             & $ \begin{array}{ccc} IRB \times3 \\ DCB \times3 \\ Channels = 224 \end{array} $          
 & $ \begin{array}{ccc} IRB \times3 \\ DCB \times3 \\ Channels = 320 \end{array} $  & $ \begin{array}{ccc} IRB \times3 \\ DCB \times3 \\ Channels = 416 \end{array} $           \\[8pt] \hline 
Head                              & 1 $\times$ 1                & Pooling \& Lin. MLP  & Pooling \& Lin. MLP            & Pooling \& Lin. MLP            & Pooling \& Lin. MLP            \\ \hline
\end{tabular}
\label{table_of_arch}
\end{table*}

The RapidNet architecture shown in Figure \ref{fig:RapidNet_Architecture}a is composed of a convolutional stem and four stages, where processing occurs at a single resolution in each stage. The stem downsamples the input image by 4$\times$ using 3 $\times$ 3 convolutions with a stride of 2 as shown in Figure \ref{fig:RapidNet_Architecture}b. The output of the stem is passed to Stage 1, which consists of  $N1i$ modified inverted residual blocks (IRB) as shown in Figure \ref{fig:RapidNet_Architecture}a. After each stage is another downsample consisting of a 3 $\times$ 3 convolution with a stride of 2 to half the input resolution and expand the channel dimension as shown in Figure \ref{fig:RapidNet_Architecture}d. Stage 2 consists of $N2i$ IRB blocks and has different channel dimensions from Stage 1. Stages 3 and 4 start with a sequence of $N3i$ and $N4i$ IRB blocks followed by $N3d$ and $N4d$ Dilated Convolution Blocks.

The IRB block is used for local processing at each stage and uses an expansion ratio of four following the method of MobileNetv2 \cite{MobileNetv2}. Each IRB block consists of a 1 $\times$ 1 convolution, BN, GeLU, a depth-wise 3 $\times$ 3 convolution, BN, GeLU, and lastly a 1 $\times$ 1 convolution plus BN and a residual connection as shown in Figure \ref{fig:RapidNet_Architecture}c. Within the IRB blocks, we replace ReLU for GeLU following \cite{EfficientFormer, MobileViG}, which show GeLU improves performance in vision tasks. The MLDC block is used for processing at a larger theoretical receptive field to better learn global object interactions.

The Dilated Convolution Block consists of the MLDC Block and the large kernel FFN shown in Figure \ref{fig:RapidNet_Architecture}e, \ref{fig:RapidNet_Architecture}f, and \ref{fig:RapidNet_Architecture}g. The MLDC Block consists of a reparameterizable 7 $\times$ 7 depthwise convolution \cite{CPE} as in \cite{FastViT}, followed by a pointwise convolution. Then, two parallel Multi-Level Dilated Convolutions with dilation factors of 2 and 3 respectively followed by another a pointwise convolution and BN. The large kernel FFN consists of a 7 $\times$ 7 depthwise convolution and BN followed by an FFN consisting of two pointwise convolutions and BN with GeLU activation in between.

\subsection{Network Configurations}

The detailed network architectures for RapidNet-Ti, S, M, and B are provided in Table \ref{table_of_arch}. We report the output size of each stage as well as the configuration of the stem, stages, and classification head. In each stage, the number of IRB and Dilated Convolution Blocks (DCB) repeated, as well as their channel dimensions are reported. As we scale up our architecture from RapidNet-Ti to RapidNet-B, we increase the number of IRB and DCB blocks (network depth) as well as the channel dimensions (network width).

\section{Experimental Results}
\label{sec:results}

\begin{table*}
\def\arraystretch{1.175}
\caption{Results of RapidNet and other mobile architectures on ImageNet-1K classification task grouped by NPU latency of an iPhone13 Mini using the ModelBench application. \textit{Type} indicates whether the model is CNN-based, CNN-ViT-based, CNN-Pooling-based, or CNN-GNN-based. \textit{Params} lists the number of model parameters in millions. \textit{GMACs} lists the number of MACs in billions. Shaded entries indicate results obtained using RapidNet.  A (-) denotes that the model could not be profiled on the iPhone 13 Mini. * indicates methods that use knowledge distillation from RegNetY-16GF \cite{RegNetY}}
\centering
\begin{tabular}[t]{c|c|c|c|c|c|c}
\hline
\multirow{2}{*}{\textbf{Model}} & \multirow{2}{*}{\textbf{Type}}  & \multirow{2}{*}{\textbf{Resolution}}  & \multirow{2}{*}{\textbf{Params (M)}} & \multirow{2}{*}{\textbf{GMACs}} & \multirow{2}{*}{\textbf{NPU Latency (ms)}} & \multirow{2}{*}{\textbf{Top-1 (\%)}} \\
                                              &         & &     &        &  &      \\ \hline
MobileNetV2x1.0 \cite{MobileNetv2}            & CNN     & $224^2$ & 3.5 & 0.3    & 0.8  & 71.8 \\
EdgeViT-XXS \cite{pan2022edgevits}            & CNN-ViT & $224^2$ & 4.1 & 0.6    & -    & 74.4 \\
MobileViG-Ti* \cite{MobileViG}                 & CNN-GNN & $224^2$ & 5.2 & 0.7    & 0.9  & 75.7 \\
\rowcolor{Gray}
\textbf{RapidNet-Ti*}                               & \textbf{CNN} & \textbf{$224^2$} & \textbf{6.6} & \textbf{0.6}    & \textbf{0.9} & \textbf{76.3} \\ \hline

MobileNetV2x1.4 \cite{MobileNetv2}            & CNN    & $224^2$ & 6.1 & 0.6    & 1.1   & 74.7 \\
EdgeViT-XS \cite{pan2022edgevits}             & CNN-ViT & $224^2$ & 6.7 & 1.1    & -     & 77.5 \\
PoolFormer-S12 \cite{MetaFormer}              & CNN-Pooling & $224^2$ & 12.0 & 1.8    & 1.5   & 77.2 \\
MobileViG-S* \cite{MobileViG}                  & CNN-GNN & $224^2$ & 7.2 & 1.0    & 1.1   & 78.2 \\
\rowcolor{Gray}
\textbf{RapidNet-S*}                               & \textbf{CNN}  & \textbf{$224^2$}  & \textbf{9.2} & \textbf{0.9}    & \textbf{1.1} & \textbf{78.6} \\ \hline

EfficientNet-B0 \cite{tan2019efficientnet}    & CNN    & $224^2$ & 5.3  & 0.4      & 1.5     & 77.7 \\
EfficientFormer-L1* \cite{EfficientFormer}     & CNN-ViT & $224^2$ & 12.3 & 1.3      & 1.3     & 79.2 \\
PoolFormer-S24 \cite{MetaFormer}              & CNN-Pooling & $224^2$ & 21.0 & 3.4    & 2.4   & 80.3 \\
MobileViG-M* \cite{MobileViG}                  & CNN-GNN & $224^2$ & 14.0 & 1.5      & 1.5     & 80.6 \\
\rowcolor{Gray}
\textbf{RapidNet-M*}                                & \textbf{CNN} &  \textbf{{$224^2$}} & \textbf{17.3} & \textbf{1.6}      & \textbf{1.6}       & \textbf{81.0} \\ \hline

EfficientNet-B3 \cite{tan2019efficientnet}    & CNN     & $224^2$ & 12.2 & 2.0       & 4.8     & 81.6 \\
MobileViTv2-1.0 \cite{MobileViTv2}            & CNN-ViT & $224^2$ & 4.9  & 1.8       & 3.0     & 78.1 \\
MobileViTv2-2.0 \cite{MobileViTv2}            & CNN-ViT & $224^2$ & 18.5 & 7.5       & 6.3     & 82.4 \\
EfficientFormer-L3* \cite{EfficientFormer}     & CNN-ViT & $224^2$ & 31.3 & 3.9       & 2.6     & 82.4 \\
EfficientFormer-L7* \cite{EfficientFormer}     & CNN-ViT & $224^2$ & 82.1 & 10.2      & 6.5     & 83.3 \\
PoolFormer-S36 \cite{MetaFormer}              & CNN-Pooling & $224^2$ & 31.0 & 5.0    & 3.3   & 81.4 \\
PoolFormer-M36 \cite{MetaFormer}              & CNN-Pooling & $224^2$ & 56.0 & 8.8    & 5.7   & 82.1 \\
\rowcolor{Gray}
    \textbf{RapidNet-B*}                                 & \textbf{CNN}  & \textbf{{$224^2$}} & \textbf{30.5} & \textbf{3.4}       & \textbf{2.7}     & \textbf{82.8} \\ \hline
\end{tabular}
\label{tab:class}
\end{table*}

In this section, we describe our experimental setup and perform a thorough comparison between RapidNet and other mobile vision architectures. Our evaluations show that for similar or fewer parameters, GMACs, and/or latency, RapidNet has a superior performance in terms of top-1 accuracy on ImageNet-1k \cite{imagenet1k} image classification, average precision (AP) on COCO \cite{coco} object detection and instance segmentation, and mean intersection over union (mIoU) on ADE20K \cite{ADE20K} semantic segmentation.

\subsection{Image Classification}

We conduct image classification experiments on the widely used ImageNet-1K \cite{imagenet1k} dataset. The dataset contains training and validation sets of approximately 1.3M images and 50K images, respectively. We train from scratch for 300 epochs with a standard resolution of 224 $\times$ 224. We implement our RapidNet model using PyTorch 1.12.1 \cite{paszke2019pytorch} and Timm library \cite{timm}. Like other mobile architectures \cite{EfficientFormerv2, MobileViG, wang2023repvit}, we use RegNetY-16GF \cite{RegNetY} with a top-1 accuracy of 82.9\% as the teacher model for knowledge distillation. Our data augmentation pipeline includes RandAugment \cite{RandAugment}, Mixup \cite{Mixup}, Cutmix \cite{CutMix}, random erasing \cite{RandomErase}, and repeated augment \cite{RepeatedAugment}. We use the AdamW \cite{AdamW} optimizer and a learning rate of $2e^{-3}$ with a cosine annealing schedule. To measure inference latency, all models are packaged as MLModels using CoreML and profiled on an iPhone 13 Mini (iOS 16) using ModelBench \cite{mobileone2022}. We use the following ModelBench settings to profile each model: 50 inference rounds, 50 inferences per round, and a low/high trim of 10. Table \ref{tab:class} shows ImageNet-1K classification results for RapidNet and other mobile architectures.

RapidNet-Ti achieves a top-1 accuracy of 76.3\%, which is higher than MobileViG-Ti \cite{MobileViG} by 0.6\% while achieving the same inference latency of 0.9 ms with 0.1 less GMACs. RapidNet-S also outperforms MobileNetV2x1.4 by 3.9\% in terms of top-1 accuracy with the same inference latency of 1.1 ms. RapidNet-M and RapidNet-B similarly, outperform competing models for similar inference latency and/or GMACs. The success of RapidNet shows the usefulness of Multi-Level Dilated Convolutions in mobile vision architectures as they can enable theoretical receptive field expansion with lower costs than ViGs and ViTs creating high accuracy and low latency CNN-based models.

\subsection{Object Detection and Instance Segmentation}

\begin{table*}[h]
\def\arraystretch{1.1}
\caption{Results of RapidNet and other backbones on COCO object detection, COCO instance segmentation, and ADE20K semantic segmentation grouped by parameters of the backbone. $AP^{box}$ and $AP^{mask}$ scores are for object detection and instance segmentation on MS COCO 2017 \cite{coco}. $mIoU$ scores are for semantic segmentation on ADE20K \cite{ADE20K}. Shaded entries indicate results obtained using RapidNet. A (-) denotes a model that did not report these results.
}
\centering
\begin{tabular}[t]{c|c|c|c|c|c|c|c|c}
\hline
\multirow{2}{*}{\textbf{Backbone}} & \multirow{2}{*}{\textbf{Params (M)}} & \multirow{2}{*}{$AP^{box}$} & \multirow{2}{*}{$AP^{box}_{50}$} & \multirow{2}{*}{$AP^{box}_{75}$} & \multirow{2}{*}{$AP^{mask}$} & \multirow{2}{*}{$AP^{mask}_{50}$} & \multirow{2}{*}{$AP^{mask}_{75}$} & \multirow{2}{*}{$mIoU$} \\
                                              & & & & & & & & \\ \hline

ResNet18 \cite{Resnet}        & 11.7  & 34.0 & 54.0 & 36.7 & 31.2 & 51.0 & 32.7 & 32.9    \\
FastViT-SA12 \cite{FastViT}                   & 10.9 & 38.9 & 60.5 & 42.2 & 35.9 & 57.6 & 38.1 & 38.0 \\
MobileViG-M \cite{MobileViG}                  & 14.0 & 41.3 & 62.8 & 45.1 & 38.1 & 60.1 & 40.8 & - \\
EfficientFormer-L1 \cite{EfficientFormer}     & 12.3 & 37.9 & 60.3 & 41.0 & 35.4 & 57.3 & 37.3 & 38.9 \\
PoolFormer-S12 \cite{MetaFormer}   & 12.0  & 37.3 & 59.0 & 40.1 & 34.6 & 55.8 & 36.9   & 37.2        \\
\rowcolor{Gray}
RapidNet-M                                    & \textbf{17.3} & \textbf{42.0} & \textbf{63.0} & \textbf{46.1} & \textbf{38.3} & \textbf{60.3} & \textbf{41.1} & \textbf{41.5}  \\ \hline

ResNet50 \cite{Resnet}            & 25.5 & 38.0 & 58.6 & 41.4 & 34.4 & 55.1 & 36.7 & 36.7    \\
Swin-T \cite{liu2021swin}        & 29.0 & 42.2 & 64.4 & 46.2 & 39.1 & 61.6 & 42.0 & 41.5   \\
PVT-Small \cite{wang2021pyramid}         & 24.5 & 40.4 & 62.9 & 43.8 & 37.8 & 60.1 & 40.3  & 39.8   \\
FastViT-SA24 \cite{FastViT}                   & 20.6 & 42.0 & 63.5 & 45.8 & 38.0 & 60.5 & 40.5 & 41.0 \\
% MobileViG-B \cite{MobileViG}                  & 2.8 & 42.0 & 64.3 & 46.0 & 38.9 & 61.4 & 41.6 & - \\
EfficientFormer-L3 \cite{EfficientFormer}     & 31.3 & 41.4 & 63.9 & 44.7 & 38.1 & 61.0 & 40.4 & 43.5 \\
EfficientFormer-L7 \cite{EfficientFormer}    & 82.1 &  42.6 & 65.1 & 46.1 & 39.0 & 62.2 & 41.7 & 45.1       \\
PoolFormer-S24 \cite{MetaFormer}          & 21.0 & 40.1 & 62.2 & 43.4 & 37.0 & 59.1 & 39.6 & 40.3    \\
\rowcolor{Gray}
RapidNet-B                                    & \textbf{30.5} & \textbf{43.1} & \textbf{64.6} & \textbf{47.2} & \textbf{39.3} & \textbf{61.5} & \textbf{42.2} & \textbf{43.8}  \\ \hline
\end{tabular}
\label{Object_Detection_Segmentation_Results}
\end{table*}

We evaluate RapidNet on MS COCO object detection and instance segmentation tasks to verify it generalizes to downstream tasks. Following \cite{MobileViG, wang2023repvit, EfficientFormer, EfficientFormerv2}, we use RapidNet as the backbone in the Mask-RCNN framework \cite{mask_r_cnn} to conduct experiments on MS COCO 2017 \cite{coco}. The dataset contains training and validations sets of 118K and 5K images, respectively. We implement the backbone using PyTorch 1.12.1 \cite{paszke2019pytorch} and Timm library \cite{timm}. The model is initialized with ImageNet-1k pretrained weights from 300 epochs of training. We use AdamW \cite{Adam, AdamW} optimizer with an initial learning rate of 2e$^{-4}$ and train the model for 12 epochs on 8 NVIDIA RTX 6000 Ada generation GPUs with a 1333 $\times$ 800 resolution following prior work \cite{li2022next, EfficientFormer, EfficientFormerv2}.

As seen in Table \ref{Object_Detection_Segmentation_Results}, with similar model size, RapidNet outperforms ResNet \cite{Resnet}, PoolFormer \cite{MetaFormer}, EfficientFormer \cite{EfficientFormer}, MobileViG \cite{MobileViG}, Swin Transformer \cite{liu2021swin}, and PVT \cite{wang2021pyramid} in terms of either parameters or improved average precision (AP) on object detection and instance segmentation. Our RapidNet-M model gets 42.0 $AP^{box}$ and 38.3 $AP^{mask}$ on the object detection and instance segmentation tasks outperforming PoolFormer-s12 \cite{MetaFormer} by 4.7 $AP^{box}$ and 3.7 $AP^{mask}$ and FastViT-SA12 \cite{FastViT} by 3.1 $AP^{box}$ and 2.4 $AP^{mask}$. Our RapidNet-B model achieves 43.1 $AP^{box}$ and 39.3 $AP^{mask}$ outperforming EfficientFormer-L3 \cite{EfficientFormer} by 1.7 $AP^{box}$ and 1.2 $AP^{mask}$ and FastViT-SA24 \cite{FastViT} by 1.1 $AP^{box}$ and 1.3 $AP^{mask}$. The strong performance of RapidNet on object detection and instance segmentation shows the capability of Multi-Level Dilated Convolutions to help RapidNet generalize well to different vision tasks.

\subsection{Semantic Segmentation}

To verify the performance of RapidNet on the semantic segmentation task, we conduct experiments on the scene parsing dataset, ADE20k \cite{ADE20K}. The dataset contains 20K training images and 2K validation images with 150 semantic categories. Following prior work \cite{MetaFormer, EfficientFormer, EfficientFormerv2, FastViT, wang2023repvit}, we integrate RapidNet as the backbone in the Semantic FPN \cite{kirillov2019panoptic} framework. The backbone is initialized with pretrained weights on ImageNet-1K and the model is trained for 40K iterations on 8 NVIDIA RTX 6000 Ada generation GPUs. We follow the process of existing works in segmentation, using AdamW \cite{Adam, AdamW} optimizer, set the learning rate as 2 $\times$ 10$^{-4}$ with a poly decay by the power of 0.9, and image resolution of 512 $\times$ 512.

As shown in Table \ref{Object_Detection_Segmentation_Results}, RapidNet-M outperforms PoolFormer-S12 \cite{MetaFormer}, FastViT-SA12 \cite{FastViT}, and EfficientFormer-L1 \cite{EfficientFormer} by 4.3, 3.5, and 2.6 mIoU. Additionally, RapidNet-B outperforms PoolFormer-S24 \cite{MetaFormer}, FastViT-SA24 \cite{FastViT}, and PVT-Small \cite{wang2021pyramid} by 3.5, 2.8, and 4.0 mIoU. Through these results we show that with MLDC, RapidNet is better able to learn long-range object interactions compared to other mobile vision architectures.

\subsection{Ablation Studies}

Ablation studies on SVGA, pointwise convolution, and 3 $\times$ 3 convolution are included in Section \ref{Subsec:Conv} of the supplementary material. Ablation studies on CPE, LK FFN, single-level dilated convolution, and MLDC are included in Section \ref{Subsec:Dilated} of the supplementary material. Ablation studies on dilation factors and kernel sizes are included in Section \ref{Subsec:Deformable} of the supplementary material.

\vspace{-3mm}

\section{Conclusion}
\label{sec:conc}
We have proposed Multi-Level Dilated Convolutions (MLDC) as a method to design mobile CNN models with larger theoretical receptive fields while maintaining low latency. MLDC is able to process features at multiple dilation levels, in parallel, thereby allowing for processing with a larger and smaller theoretical receptive field, then combining that information to enhance feature extraction. Additionally, we have also proposed a novel CNN-based architecture, RapidNet, which uses a combination of MLDC, inverted residual blocks, a large kernel feedforward network, and reparameterizable large kernel depthwise convolutions. 

RapidNet outperforms existing CNN, ViG, ViT, and hybrid models on image classification, object detection, instance segmentation, and semantic segmentation. RapidNet performs particularly well on the downstream tasks of object detection, instance segmentation, and semantic segmentation due to the ability of dilated convolutions to better learn long-range object interactions compared to standard convolutions. The effectiveness of RapidNet shows the ability of CNN-based networks to compete with state-of-the-art ViT and hybrid CNN-ViT models.

\vspace{-2mm}

\section{Acknowledgements}
\label{Sec:Acknowledgement}

This work is supported in part by the NSF grant CNS 2007284, and in part by the iMAGiNE Consortium (\url{https://imagine.utexas.edu/}).

\newpage

%%%%%%%%% REFERENCES
{\small
\bibliographystyle{ieee_fullname}
\bibliography{egbib}

\begin{thebibliography}{10}\itemsep=-1pt

\bibitem{MobileViGv2}
William Avery, Mustafa Munir, and Radu Marculescu.
\newblock Scaling graph convolutions for mobile vision.
\newblock In {\em Proceedings of the IEEE/CVF Conference on Computer Vision and Pattern Recognition (CVPR) Workshops}, pages 5857--5865, June 2024.

\bibitem{brock2018large}
Andrew Brock, Jeff Donahue, and Karen Simonyan.
\newblock Large scale gan training for high fidelity natural image synthesis.
\newblock {\em arXiv preprint arXiv:1809.11096}, 2018.

\bibitem{carion2020end}
Nicolas Carion, Francisco Massa, Gabriel Synnaeve, Nicolas Usunier, Alexander Kirillov, and Sergey Zagoruyko.
\newblock End-to-end object detection with transformers.
\newblock In {\em Proceedings of the European Conference on Computer Vision}, pages 213--229. Springer, 2020.

\bibitem{chen2017deeplab}
Liang-Chieh Chen, George Papandreou, Iasonas Kokkinos, Kevin Murphy, and Alan~L Yuille.
\newblock Deeplab: Semantic image segmentation with deep convolutional nets, atrous convolution, and fully connected crfs.
\newblock {\em IEEE transactions on pattern analysis and machine intelligence}, 40(4):834--848, 2017.

\bibitem{MobileFormer}
Yinpeng Chen, Xiyang Dai, Dongdong Chen, Mengchen Liu, Xiaoyi Dong, Lu Yuan, and Zicheng Liu.
\newblock Mobile-former: Bridging mobilenet and transformer.
\newblock In {\em Proceedings of the IEEE/CVF Conference on Computer Vision and Pattern Recognition}, pages 5270--5279, 2022.

\bibitem{CPE}
Xiangxiang Chu, Zhi Tian, Bo Zhang, Xinlong Wang, Xiaolin Wei, Huaxia Xia, and Chunhua Shen.
\newblock Conditional positional encodings for vision transformers.
\newblock {\em arXiv preprint arXiv:2102.10882}, 2021.

\bibitem{RandAugment}
Ekin~D Cubuk, Barret Zoph, Jonathon Shlens, and Quoc~V Le.
\newblock Randaugment: Practical automated data augmentation with a reduced search space.
\newblock In {\em Proceedings of the IEEE/CVF Conference on Computer Vision and Pattern Recognition Workshops}, pages 702--703, 2020.

\bibitem{dai2017deformable}
Jifeng Dai, Haozhi Qi, Yuwen Xiong, Yi Li, Guodong Zhang, Han Hu, and Yichen Wei.
\newblock Deformable convolutional networks.
\newblock In {\em Proceedings of the IEEE international conference on computer vision}, pages 764--773, 2017.

\bibitem{imagenet1k}
Jia Deng, Wei Dong, Richard Socher, Li-Jia Li, Kai Li, and Li Fei-Fei.
\newblock Imagenet: A large-scale hierarchical image database.
\newblock In {\em 2009 IEEE Conference on Computer Vision and Pattern Recognition}, pages 248--255, 2009.

\bibitem{dhariwal2021diffusion}
Prafulla Dhariwal and Alexander Nichol.
\newblock Diffusion models beat gans on image synthesis.
\newblock {\em Advances in Neural Information Processing Systems}, 34:8780--8794, 2021.

\bibitem{31_31_Conv}
Xiaohan Ding, Xiangyu Zhang, Jungong Han, and Guiguang Ding.
\newblock Scaling up your kernels to 31x31: Revisiting large kernel design in cnns.
\newblock In {\em Proceedings of the IEEE/CVF Conference on Computer Vision and Pattern Recognition}, pages 11963--11975, 2022.

\bibitem{ding2021repvgg}
Xiaohan Ding, Xiangyu Zhang, Ningning Ma, Jungong Han, Guiguang Ding, and Jian Sun.
\newblock Repvgg: Making vgg-style convnets great again.
\newblock In {\em Proceedings of the IEEE/CVF Conference on Computer Vision and Pattern Recognition}, pages 13733--13742, 2021.

\bibitem{ViT}
Alexey Dosovitskiy et~al.
\newblock An image is worth 16x16 words: Transformers for image recognition at scale.
\newblock {\em arXiv preprint arXiv:2010.11929}, 2020.

\bibitem{esser2023structure}
Patrick Esser, Johnathan Chiu, Parmida Atighehchian, Jonathan Granskog, and Anastasis Germanidis.
\newblock Structure and content-guided video synthesis with diffusion models.
\newblock In {\em Proceedings of the IEEE/CVF International Conference on Computer Vision}, pages 7346--7356, 2023.

\bibitem{goodfellow2014generative}
Ian Goodfellow, Jean Pouget-Abadie, Mehdi Mirza, Bing Xu, David Warde-Farley, Sherjil Ozair, Aaron Courville, and Yoshua Bengio.
\newblock Generative adversarial nets.
\newblock {\em Advances in Neural Information Processing Systems}, 27, 2014.

\bibitem{Vision_GNN}
Kai Han, Yunhe Wang, Jianyuan Guo, Yehui Tang, and Enhua Wu.
\newblock Vision gnn: An image is worth graph of nodes.
\newblock {\em arXiv preprint arXiv:2206.00272}, 2022.

\bibitem{ViHGNN}
Yan Han, Peihao Wang, Souvik Kundu, Ying Ding, and Zhangyang Wang.
\newblock Vision hgnn: An image is more than a graph of nodes.
\newblock In {\em Proceedings of the IEEE/CVF International Conference on Computer Vision}, pages 19878--19888, 2023.

\bibitem{mask_r_cnn}
Kaiming He, Georgia Gkioxari, Piotr Doll{\'a}r, and Ross Girshick.
\newblock Mask r-cnn.
\newblock In {\em Proceedings of the IEEE International Conference on Computer Vision}, pages 2961--2969, 2017.

\bibitem{Resnet}
Kaiming He, Xiangyu Zhang, Shaoqing Ren, and Jian Sun.
\newblock Deep residual learning for image recognition.
\newblock In {\em Proceedings of the IEEE/CVF Conference on Computer Vision and Pattern Recognition}, pages 770--778, 2016.

\bibitem{GeLU}
Dan Hendrycks and Kevin Gimpel.
\newblock Gaussian error linear units (gelus).
\newblock {\em arXiv preprint arXiv:1606.08415}, 2016.

\bibitem{ho2020denoising}
Jonathan Ho, Ajay Jain, and Pieter Abbeel.
\newblock Denoising diffusion probabilistic models.
\newblock {\em Advances in Neural Information Processing Systems}, 33:6840--6851, 2020.

\bibitem{RepeatedAugment}
Elad Hoffer, Tal Ben-Nun, Itay Hubara, Niv Giladi, Torsten Hoefler, and Daniel Soudry.
\newblock Augment your batch: Improving generalization through instance repetition.
\newblock In {\em Proceedings of the IEEE/CVF Conference on Computer Vision and Pattern Recognition}, pages 8129--8138, 2020.

\bibitem{holschneider1990real}
Matthias Holschneider, Richard Kronland-Martinet, Jean Morlet, and Ph Tchamitchian.
\newblock A real-time algorithm for signal analysis with the help of the wavelet transform.
\newblock In {\em Wavelets: Time-Frequency Methods and Phase Space Proceedings of the International Conference, Marseille, France, December 14--18, 1987}, pages 286--297. Springer, 1990.

\bibitem{MobileNet}
Andrew~G Howard, Menglong Zhu, Bo Chen, Dmitry Kalenichenko, Weijun Wang, Tobias Weyand, Marco Andreetto, and Hartwig Adam.
\newblock Mobilenets: Efficient convolutional neural networks for mobile vision applications.
\newblock {\em arXiv preprint arXiv:1704.04861}, 2017.

\bibitem{Densenet}
Gao Huang, Zhuang Liu, Laurens Van Der~Maaten, and Kilian~Q Weinberger.
\newblock Densely connected convolutional networks.
\newblock In {\em Proceedings of the IEEE/CVF Conference on Computer Vision and Pattern Recognition}, pages 4700--4708, 2017.

\bibitem{huang2018imsynthesis}
He Huang, Philip~S Yu, and Changhu Wang.
\newblock An introduction to image synthesis with generative adversarial nets.
\newblock {\em arXiv preprint arXiv:1803.04469}, 2018.

\bibitem{SqueezeNet}
Forrest~N. Iandola, Song Han, Matthew~W. Moskewicz, Khalid Ashraf, William~J. Dally, and Kurt Keutzer.
\newblock Squeezenet: Alexnet-level accuracy with 50x fewer parameters and $<$0.5mb model size.
\newblock {\em arXiv:1602.07360}, 2016.

\bibitem{BatchNorm}
Sergey Ioffe and Christian Szegedy.
\newblock Batch normalization: Accelerating deep network training by reducing internal covariate shift.
\newblock In {\em International Conference on Machine Learning}, pages 448--456. pmlr, 2015.

\bibitem{Adam}
Diederik~P Kingma and Jimmy Ba.
\newblock Adam: A method for stochastic optimization.
\newblock {\em arXiv preprint arXiv:1412.6980}, 2014.

\bibitem{kirillov2019panoptic}
Alexander Kirillov, Ross Girshick, Kaiming He, and Piotr Doll{\'a}r.
\newblock Panoptic feature pyramid networks.
\newblock In {\em Proceedings of the IEEE/CVF Conference on Computer Vision and Pattern Recognition}, pages 6399--6408, 2019.

\bibitem{lecun1998gradient}
Yann LeCun, L{\'e}on Bottou, Yoshua Bengio, and Patrick Haffner.
\newblock Gradient-based learning applied to document recognition.
\newblock {\em Proceedings of the IEEE}, 86(11):2278--2324, 1998.

\bibitem{maxrel}
Guohao Li, Matthias Muller, Ali Thabet, and Bernard Ghanem.
\newblock Deepgcns: Can gcns go as deep as cnns?
\newblock In {\em Proceedings of the IEEE/CVF International Conference on Computer Vision}, pages 9267--9276, 2019.

\bibitem{li2022next}
Jiashi Li, Xin Xia, Wei Li, Huixia Li, Xing Wang, Xuefeng Xiao, Rui Wang, Min Zheng, and Xin Pan.
\newblock Next-vit: Next generation vision transformer for efficient deployment in realistic industrial scenarios.
\newblock {\em arXiv preprint arXiv:2207.05501}, 2022.

\bibitem{EfficientFormerv2}
Yanyu Li, Ju Hu, Yang Wen, Georgios Evangelidis, Kamyar Salahi, Yanzhi Wang, Sergey Tulyakov, and Jian Ren.
\newblock Rethinking vision transformers for mobilenet size and speed.
\newblock {\em arXiv preprint arXiv:2212.08059}, 2022.

\bibitem{EfficientFormer}
Yanyu Li, Geng Yuan, Yang Wen, Eric Hu, Georgios Evangelidis, Sergey Tulyakov, Yanzhi Wang, and Jian Ren.
\newblock Efficientformer: Vision transformers at mobilenet speed.
\newblock {\em arXiv preprint arXiv:2206.01191}, 2022.

\bibitem{coco}
Tsung-Yi Lin, Michael Maire, Serge Belongie, James Hays, Pietro Perona, Deva Ramanan, Piotr Doll{\'a}r, and C~Lawrence Zitnick.
\newblock Microsoft coco: Common objects in context.
\newblock In {\em Proceedings of the European Conference on Computer Vision}, pages 740--755. Springer, 2014.

\bibitem{liu2021generative}
Ming-Yu Liu, Xun Huang, Jiahui Yu, Ting-Chun Wang, and Arun Mallya.
\newblock Generative adversarial networks for image and video synthesis: Algorithms and applications.
\newblock {\em Proceedings of the IEEE}, 109(5):839--862, 2021.

\bibitem{liu2021swin}
Ze Liu, Yutong Lin, Yue Cao, Han Hu, Yixuan Wei, Zheng Zhang, Stephen Lin, and Baining Guo.
\newblock Swin transformer: Hierarchical vision transformer using shifted windows.
\newblock In {\em Proceedings of the IEEE/CVF International Conference on Computer Vision}, pages 10012--10022, 2021.

\bibitem{liu2022convnet}
Zhuang Liu, Hanzi Mao, Chao-Yuan Wu, Christoph Feichtenhofer, Trevor Darrell, and Saining Xie.
\newblock A convnet for the 2020s.
\newblock In {\em Proceedings of the IEEE/CVF Conference on Computer Vision and Pattern Recognition}, pages 11976--11986, 2022.

\bibitem{AdamW}
Ilya Loshchilov and Frank Hutter.
\newblock Decoupled weight decay regularization.
\newblock {\em arXiv preprint arXiv:1711.05101}, 2017.

\bibitem{shufflenetv2}
Ningning Ma, Xiangyu Zhang, Hai-Tao Zheng, and Jian Sun.
\newblock Shufflenet v2: Practical guidelines for efficient cnn architecture design.
\newblock In {\em Proceedings of the European conference on computer vision (ECCV)}, pages 116--131, 2018.

\bibitem{mao2022towards}
Xiaofeng Mao, Gege Qi, Yuefeng Chen, Xiaodan Li, Ranjie Duan, Shaokai Ye, Yuan He, and Hui Xue.
\newblock Towards robust vision transformer.
\newblock In {\em Proceedings of the IEEE/CVF Conference on Computer Vision and Pattern Recognition}, pages 12042--12051, 2022.

\bibitem{MobileViT}
Sachin Mehta and Mohammad Rastegari.
\newblock Mobilevit: light-weight, general-purpose, and mobile-friendly vision transformer.
\newblock {\em arXiv preprint arXiv:2110.02178}, 2021.

\bibitem{MobileViTv2}
Sachin Mehta and Mohammad Rastegari.
\newblock Separable self-attention for mobile vision transformers.
\newblock {\em arXiv preprint arXiv:2206.02680}, 2022.

\bibitem{mehta2018espnet}
Sachin Mehta, Mohammad Rastegari, Anat Caspi, Linda Shapiro, and Hannaneh Hajishirzi.
\newblock Espnet: Efficient spatial pyramid of dilated convolutions for semantic segmentation.
\newblock In {\em Proceedings of the European Conference on Computer Vision}, pages 552--568, 2018.

\bibitem{MobileViG}
Mustafa Munir, William Avery, and Radu Marculescu.
\newblock Mobilevig: Graph-based sparse attention for mobile vision applications.
\newblock In {\em Proceedings of the IEEE/CVF Conference on Computer Vision and Pattern Recognition Workshops}, pages 2211--2219, 2023.

\bibitem{GreedyViG}
Mustafa Munir, William Avery, Md~Mostafijur Rahman, and Radu Marculescu.
\newblock Greedyvig: Dynamic axial graph construction for efficient vision gnns.
\newblock In {\em Proceedings of the IEEE/CVF Conference on Computer Vision and Pattern Recognition (CVPR)}, pages 6118--6127, June 2024.

\bibitem{munir2024three}
Mustafa Munir, Saloni Modi, Geffen Cooper, HunTae Kim, and Radu Marculescu.
\newblock Three decades of low power: From watts to wisdom.
\newblock {\em IEEE Access}, 2024.

\bibitem{pan2022edgevits}
Junting Pan, Adrian Bulat, Fuwen Tan, Xiatian Zhu, Lukasz Dudziak, Hongsheng Li, Georgios Tzimiropoulos, and Brais Martinez.
\newblock Edgevits: Competing light-weight cnns on mobile devices with vision transformers.
\newblock In {\em Proceedings of the European Conference on Computer Vision}, pages 294--311. Springer, 2022.

\bibitem{paszke2019pytorch}
Adam Paszke et~al.
\newblock Pytorch: An imperative style, high-performance deep learning library.
\newblock {\em Advances in Neural Information Processing Systems}, 32, 2019.

\bibitem{RegNetY}
Ilija Radosavovic, Raj~Prateek Kosaraju, Ross Girshick, Kaiming He, and Piotr Doll{\'a}r.
\newblock Designing network design spaces.
\newblock In {\em Proceedings of the IEEE/CVF Conference on Computer Vision and Pattern Recognition}, pages 10428--10436, 2020.

\bibitem{MobileNetv2}
Mark Sandler, Andrew Howard, Menglong Zhu, Andrey Zhmoginov, and Liang-Chieh Chen.
\newblock Mobilenetv2: Inverted residuals and linear bottlenecks.
\newblock In {\em Proceedings of the IEEE conference on computer vision and pattern recognition}, pages 4510--4520, 2018.

\bibitem{tan2019efficientnet}
Mingxing Tan and Quoc Le.
\newblock Efficientnet: Rethinking model scaling for convolutional neural networks.
\newblock In {\em International Conference on Machine Learning}, pages 6105--6114. PMLR, 2019.

\bibitem{tan2021efficientnetv2}
Mingxing Tan and Quoc Le.
\newblock Efficientnetv2: Smaller models and faster training.
\newblock In {\em International Conference on Machine Learning}, pages 10096--10106. PMLR, 2021.

\bibitem{mlpmixer}
Ilya~O Tolstikhin, Neil Houlsby, Alexander Kolesnikov, Lucas Beyer, Xiaohua Zhai, Thomas Unterthiner, Jessica Yung, Andreas Steiner, Daniel Keysers, Jakob Uszkoreit, et~al.
\newblock Mlp-mixer: An all-mlp architecture for vision.
\newblock {\em Advances in Neural Information Processing Systems}, 34:24261--24272, 2021.

\bibitem{resmlp}
Hugo Touvron, Piotr Bojanowski, Mathilde Caron, Matthieu Cord, Alaaeldin El-Nouby, Edouard Grave, Gautier Izacard, Armand Joulin, Gabriel Synnaeve, Jakob Verbeek, et~al.
\newblock Resmlp: Feedforward networks for image classification with data-efficient training.
\newblock {\em IEEE Transactions on Pattern Analysis and Machine Intelligence}, 2022.

\bibitem{mobileone2022}
Pavan Kumar~Anasosalu Vasu, James Gabriel, Jeff Zhu, Oncel Tuzel, and Anurag Ranjan.
\newblock An improved one millisecond mobile backbone.
\newblock {\em arXiv preprint arXiv:2206.04040}, 2022.

\bibitem{FastViT}
Pavan Kumar~Anasosalu Vasu, James Gabriel, Jeff Zhu, Oncel Tuzel, and Anurag Ranjan.
\newblock Fastvit: A fast hybrid vision transformer using structural reparameterization.
\newblock In {\em Proceedings of the IEEE/CVF International Conference on Computer Vision}, 2023.

\bibitem{vaswani2017attention}
Ashish Vaswani, Noam Shazeer, Niki Parmar, Jakob Uszkoreit, Llion Jones, Aidan~N Gomez, {\L}ukasz Kaiser, and Illia Polosukhin.
\newblock Attention is all you need.
\newblock {\em Advances in Neural Information Processing Systems}, 30, 2017.

\bibitem{wang2023repvit}
Ao Wang, Hui Chen, Zijia Lin, Hengjun Pu, and Guiguang Ding.
\newblock Repvit: Revisiting mobile cnn from vit perspective.
\newblock {\em arXiv preprint arXiv:2307.09283}, 2023.

\bibitem{wang2021pyramid}
Wenhai Wang, Enze Xie, Xiang Li, Deng-Ping Fan, Kaitao Song, Ding Liang, Tong Lu, Ping Luo, and Ling Shao.
\newblock Pyramid vision transformer: A versatile backbone for dense prediction without convolutions.
\newblock In {\em Proceedings of the IEEE/CVF International Conference on Computer Vision}, pages 568--578, 2021.

\bibitem{wang2022can}
Zeyu Wang, Yutong Bai, Yuyin Zhou, and Cihang Xie.
\newblock Can cnns be more robust than transformers?
\newblock {\em arXiv preprint arXiv:2206.03452}, 2022.

\bibitem{timm}
Ross Wightman.
\newblock {PyTorch Image Models}.
\newblock \url{https://github.com/rwightman/pytorch-image-models}, 2019.

\bibitem{yu2015multi}
Fisher Yu and Vladlen Koltun.
\newblock Multi-scale context aggregation by dilated convolutions.
\newblock {\em arXiv preprint arXiv:1511.07122}, 2015.

\bibitem{MetaFormer}
Weihao Yu, Mi Luo, Pan Zhou, Chenyang Si, Yichen Zhou, Xinchao Wang, Jiashi Feng, and Shuicheng Yan.
\newblock Metaformer is actually what you need for vision.
\newblock In {\em Proceedings of the IEEE/CVF Conference on Computer Vision and Pattern Recognition}, pages 10819--10829, 2022.

\bibitem{CutMix}
Sangdoo Yun, Dongyoon Han, Seong~Joon Oh, Sanghyuk Chun, Junsuk Choe, and Youngjoon Yoo.
\newblock Cutmix: Regularization strategy to train strong classifiers with localizable features.
\newblock In {\em Proceedings of the IEEE/CVF International Conference on Computer Vision}, pages 6023--6032, 2019.

\bibitem{Mixup}
Hongyi Zhang, Moustapha Cisse, Yann~N. Dauphin, and David Lopez-Paz.
\newblock mixup: Beyond empirical risk minimization.
\newblock In {\em International Conference on Learning Representations}, 2018.

\bibitem{shufflenet}
Xiangyu Zhang, Xinyu Zhou, Mengxiao Lin, and Jian Sun.
\newblock Shufflenet: An extremely efficient convolutional neural network for mobile devices.
\newblock In {\em Proceedings of the IEEE conference on computer vision and pattern recognition}, pages 6848--6856, 2018.

\bibitem{RandomErase}
Zhun Zhong, Liang Zheng, Guoliang Kang, Shaozi Li, and Yi Yang.
\newblock Random erasing data augmentation.
\newblock In {\em Proceedings of the AAAI conference on artificial intelligence}, volume~34, pages 13001--13008, 2020.

\bibitem{ADE20K}
Bolei Zhou, Hang Zhao, Xavier Puig, Sanja Fidler, Adela Barriuso, and Antonio Torralba.
\newblock Scene parsing through ade20k dataset.
\newblock In {\em Proceedings of the IEEE/CVF Conference on Computer Vision and Pattern Recognition}, pages 633--641, 2017.

\end{thebibliography}
}

%% redefine the \title command so that a variable name is saved in \thetitle, and provides the \maketitlesupplementary command 
\let\titleold\title
\def\maketitlesupplementary
   {
   \newpage
       \twocolumn[
        \centering
        \Large
        \textbf{RapidNet: Multi-Level Dilated Convolution Based Mobile Backbone}\\
        \vspace{0.5em}Supplementary Material \\
        \vspace{1.0em}
       ] %< twocolumn
   }

\clearpage
\setcounter{page}{1}
\maketitlesupplementary
\appendix

\section{Ablation Studies}
\label{Sec:Ablations}

The ablation studies are conducted on ImageNet-1K \cite{imagenet1k}. Table \ref{tab:ablation_conv} reports the ablation study of RapidNet-Ti (RNet-Ti) on the effects of static graph convolution, pointwise convolution, and 3 $\times$ 3 convolution. Table \ref{tab:ablation_RNet} reports the effects of conditional positional encoding (CPE), the large kernel feedforward network (FFN), single-level dilated convolution (SLDC), and multi-level dilated convolution (MLDC).

\subsection{Effect of Different Convolution Types and Knowledge Distillation}
\label{Subsec:Conv}

Starting with a RapidNet-Ti configuration with no CPE, no large kernel FFN, and using pointwise convolution instead of MLDC we achieve a top-1 accuracy of 75.2\%. We can see this is a lower accuracy than the static graph convolution of MobileViG-Ti in Table \ref{tab:ablation_conv}, which achieves 75.7\% with an increase of 0.3 GMACs (42.9\% increase in GMACs). This shows that static graph convolution adds additional information beyond pointwise convolution. When we replace the PW convolution with a 3$\times$3 kernel convolution we gain 0.5\% in accuracy compared to the PW convolution, but we also gain 0.1 GMACs. This demonstrates that our RapidNet architecture can match MobileViG in accuracy with a lower amount of GMACs by simply using 3 $\times$ 3 kernel convolutions in our architecture.

\begin{table}[h]
\footnotesize
\def\arraystretch{1.3}
\setlength{\tabcolsep}{2pt} % Adjust column separation
\caption{Ablation study for RapidNet-Ti on ImageNet-1K benchmark for how SVGA \cite{MobileViG}, pointwise (PW) convolution, 3$\times$3 kernel convolution, and knowledge distillation (KD) affect performance. In each column yes means this form of convolution was used in the network in place of multi-level dilated convolution in the dilated convolution block. No means this form of convolution was not used to replace multi-level dilated convolution in the ablation study. For KD yes and no just mean whether KD was used. For the MLDC column yes means MLDC was used and no replacement method of convolution was used.}
\centering
\begin{tabular}[t]{|c|c|c|c|c|c|c|c|c|c|}
\hline
\textbf{Model} & \textbf{GMACs} & \textbf{SVGA} & \textbf{MLDC}& \textbf{PW Conv} & \textbf{3$\times$3 Conv}  & \textbf{KD} & \textbf{Acc}  \\ \hline
MViG-Ti & 0.7   & Yes & No    & Yes & No      & Yes  & 75.7 \\ \hline
RNet-Ti & 0.4    & No  & No  & Yes & No   & Yes     & 75.2 \\ \hline
RNet-Ti & 0.5    & No  & No & No  &  Yes    & Yes  & 75.7 \\ \hline
MViG-Ti & 0.7    & Yes & No & No  &  No    & No  & 74.5 \\ \hline
RNet-Ti & 0.6    & No & Yes & No  &  No    & No  & 75.1 \\ \hline
\rowcolor {Gray}
\textbf{RNet-Ti} & \textbf{0.6}    & \textbf{No}  & \textbf{Yes} & \textbf{No}  &  \textbf{No}    & \textbf{Yes}  & \textbf{76.3} \\ \hline
\end{tabular}
\label{tab:ablation_conv}
\end{table}

When trying to determine the impact of knowledge distillation in Table \ref{tab:ablation_conv} we can see RapidNet-Ti loses 1.2\% in accuracy as does MobileViG-Ti. This shows that for both models knowledge distillation is beneficial to performance.

\subsection{Effect of Dilated Convolutions and Positional Encoding}
\label{Subsec:Dilated}

Replacing the 3$\times$3 kernel convolution in Table \ref{tab:ablation_conv} with a single-level dilated convolution in Table \ref{tab:ablation_RNet}, we gain another 0.1\% increase in accuracy with no increase in the number of GMACs. Adding CPE and the large kernel FFN increases the accuracy by 0.1\% each with a near negligible gain in GMACs. Lastly adding MLDC to replace SLDC increases the accuracy to 76.3\% providing another 0.3\% increase with an increase of only 0.1 GMACs.

\begin{table}[h]
\footnotesize
\def\arraystretch{1.2}
\setlength{\tabcolsep}{4.5pt} % Adjust column separation
\caption{Ablation study for RapidNet-Ti on ImageNet-1K benchmark for how CPE, large kernel feedforward network (LKFFN), single-level dilated convolution (SLDC),  and multi-level dilated convolution (MLDC) affect performance. Results are averaged over two runs.}
\centering
\begin{tabular}[t]{|c|c|c|c|c|c|c|c|c|c|}
\hline
\textbf{Model} & \textbf{GMACs} &  \textbf{CPE} &  \textbf{ LKFFN} & \textbf{SLDC} & \textbf{MLDC} & \textbf{Acc (\%)}  \\ \hline
RNet-Ti & 0.5    & No   & No  &  Yes & No        & 75.8 \\ \hline
RNet-Ti & 0.5     & Yes   & No  &  Yes & No        & 75.9 \\ \hline
RNet-Ti & 0.5     & Yes   & Yes  &  Yes & No        & 76.0 \\ \hline
\rowcolor {Gray}
\textbf{RNet-Ti} & \textbf{0.6} & \textbf{Yes}  &    \textbf{Yes}  & \textbf{No}  & \textbf{Yes}        &  \textbf{76.3}  \\ \hlineB{5}
\end{tabular}
\label{tab:ablation_RNet}
\end{table}

\subsection{Effect of Dilation Factors, Kernel Sizes, and Deformable Convolution}
\label{Subsec:Deformable}

Replacing the 3$\times$3 kernel convolution in MLDC with 5$\times$5 kernel convolution in Table \ref{tab:ablation_kernel_dilation_deformable}, we gain only 0.1\% in accuracy, but we gain 2 million parameters. Due to this increased computational cost and minimal benefit in terms of accuracy we opt for 3$\times$3 kernel convolution in our network. We also perform an ablation study using larger dilation factors in our RapidNet model by increasing the dilation factor in MLDC from 2 and 3 to 3 and 4. Increasing the dilation factor actually leads to a decrease in accuracy of 0.4\% falling from 76.3\% to 75.9\%. Replacing MLDC with deformable convolution we gain 0.1 million parameters due to the learnable offsets, but we do not see an increase in accuracy. Instead we actually see a decrease in accuracy of 0.4\%. For image classification tasks, dilated convolutions are more widely used as opposed to deformable convolutions which are more widely used for tasks like object detection and segmentation \cite{chen2017deeplab, dai2017deformable}.

\begin{table}[h]
\footnotesize
\def\arraystretch{1.2}
\setlength{\tabcolsep}{3.3pt} % Adjust column separation
\caption{Ablation study for RapidNet-Ti on ImageNet-1K for how kernel size and dilation factors in MLDC affect performance.}
\centering
\begin{tabular}[t]{|c|c|c|c|c|c|c|c|c|c|}
\hline
\textbf{Model} & \textbf{Params (M)} &  \textbf{Kernel} &  \textbf{Dilation} & \textbf{Deformable} & \textbf{Acc (\%)}  \\ \hline
RNet-Ti & 6.7    & 3x3   & No  &  Yes       & 76.0 \\ \hline
RNet-Ti & 6.6     & 3x3   & 3,4  &  No       & 75.9 \\ \hline
RNet-Ti & 8.6     & 5x5   & 2,3  &  No       & 76.4 \\ \hline
\rowcolor {Gray}
\textbf{RNet-Ti} & \textbf{6.6} & \textbf{3x3}  &    \textbf{2,3}  & \textbf{No}   &  \textbf{76.3}  \\ \hlineB{5}
\end{tabular}
\label{tab:ablation_kernel_dilation_deformable}
\end{table}

\section{Further Latency Results}
\label{Sec:Latencies}
\subsection{RapidNet ImageNet-1k Classification versus Latency}
\label{Subsec:Latency}

\begin{figure}[h]
\centering
\includegraphics[width=\linewidth,]{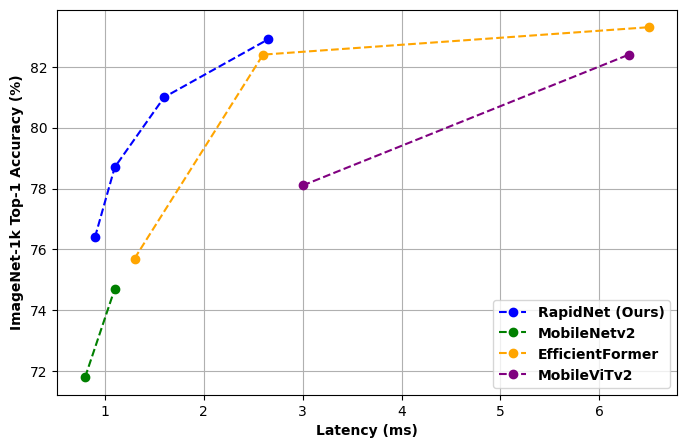}
\caption{\textbf{Comparison of accuracy versus latency on ImageNet-1K}. RapidNet achieves the best accuracy-latency tradeoff on all model sizes compared.}
\label{fig:pareto_latency}
\end{figure}

We visualize RapidNet's accuracy-latency tradeoff in Figure \ref{fig:pareto_latency} to demonstrate we are not only optimal in terms of the accuracy-GMACs tradeoff as shown in Figure \ref{fig:pareto}, but also in terms of accuracy versus latency.

\end{document}